\documentclass[twoside]{BioRxiv}

\usepackage{graphicx}
\usepackage{epstopdf}
\usepackage{booktabs}
\usepackage{enumitem}
\usepackage{verbatim}
\usepackage{dblfloatfix}

\begin{document}

\title{Deep learning tools for the measurement of animal
behavior in neuroscience}
\shorttitle{deep learning \& animal behavior}
\author[1,*]{Mackenzie Weygandt Mathis}
\author[1]{Alexander Mathis}
\leadauthor{Mathis \& Mathis}
\affil[1]{Harvard University, Cambridge, MA USA | *mackenzie@post.harvard.edu} 
\maketitle

\begin{abstract}
Recent advances in computer vision have made accurate, fast and robust measurement of animal behavior a reality. In the past years powerful tools specifically designed to aid the measurement of behavior have come to fruition. Here we discuss how capturing the postures of animals - pose estimation - has been rapidly advancing with new deep learning methods. While challenges still remain, we envision that the fast-paced development of new deep learning tools will rapidly change the landscape of realizable real-world neuroscience. 
\end{abstract}
{\bf Highlights:}
\begin{enumerate}
\item Deep neural networks are shattering performance benchmarks in computer vision for various tasks.
\item Using modern deep learning approaches (DNNs) in the lab is a fruitful approach for robust, fast, and efficient measurement of animal behavior.
\item New DNN-based tools allow for customized tracking approaches, which opens new avenues for more flexible and ethologically relevant real-world neuroscience.
\end{enumerate}

\section*{Introduction}
Behavior is the most important output of the underlying neural computations in the brain. Behavior is complex, often multi-faceted, and highly context dependent both in how con-specifics or other observers understand it, as well as how it is emitted. The study of animal behavior - ethology - has a rich history rooted in the understanding that behavior gives an observer a unique look into an animal's \textit{umwelt} ~\cite{Uexkull, Tinbergen1963, Bernstein1967}; what are the motivations, instincts, and needs of an animal? What survival value do they provide? In order to understand the brain, we need to measure behavior in all its beauty and depth, and distill it down into meaningful metrics. Observing and efficiently describing behavior is a core tenant of modern ethology, neuroscience, medicine, and technology.\\

In 1973 Tinbergen, Lorenz, and von Frisch were the first ethologists awarded the Nobel Prize in Physiology or Medicine for their pioneering work on the patterns of individual and social group behavior~\cite{tinbergen73}. The award heralded a coming-of-age for behavior, and how rigorously documenting behavior can influence how we study the brain~\cite{tinbergen73}. Manual methods are powerful, but also highly labor intensive and subject to the limits of our senses. Matching (and extending) the capabilities of biologists with technology is a highly non-trivial problem~\cite{Schaefer2012TheSS, Anderson2014}, yet harbors tremendous potential. How does one compress an animal's behavior over long time periods into meaningful metrics?  How does one use behavioral quantification to build a better understanding of the brain and an animal's \textit{umwelt}~\cite{Uexkull}?\\

In this review we discuss the advances, and challenges, in animal pose estimation and its impact on neuroscience. Pose estimation refers to methods for measuring posture, while posture denotes to the geometrical configuration of body parts. While there are many ways to record behavior~\cite{Dell2014review, egnor2016computational, camomilla2018trends}, videography is a non-invasive way to observe the posture of animals. Estimated poses across time can then, depending on the application, be transformed into kinematics, dynamics, and actions~\cite{Bernstein1967,gomez2014big, Schaefer2012TheSS, Dell2014review, Anderson2014}. Due to the low-dimensional nature of posture, these applications are computationally tractable.

\subsection*{A very brief history of pose estimation}

\justify The postures and actions of animals have been documented as far back as cave paintings, illustrating the human desire to distill the essence of an animal for conveying information. As soon as it was possible to store data on a computer, researchers have built systems for automated analysis. Over time, these systems reflected all flavors of artificial intelligence from rule-based via expert systems, to machine learning~\cite{POPPE20074,litjens2017survey}. Traditionally posture was measured by placing markers on the subject~\cite{Bernstein1967}, or markerlessly by using body models (i.e. cylinder-based models with edge features~\cite{HOGG19835}). Other computer vision techniques, such as using texture or color to segment the person from the background to create silhouettes~\cite{wren1997pfinder,cremers2007review}, or using so-called hand-crafted features with decoders~\cite{POPPE20074, litjens2017survey,serre2019deep} were also popular before deep learning flourished.

\subsection*{The deep learning revolution for posture} 

\justify Pose estimation is a challenging problem, but it has been tremendously advanced in the last five years due to advances in deep learning. Deep neural networks (DNNs) are computational algorithms that consist of simple units, which are  organized in layers and then serially stacked to form ``deep networks". The connections between the units are trained on data and therefore learn to extract information from raw data in order to solve tasks. The current deep learning revolution started with achieving human-level accuracy for object recognition on the ImageNet challenge, a popular benchmark with many categories and millions of images~\cite{alom2018history, serre2019deep}. A combination of large annotated data sets, sophisticated network architectures, and advances in hardware made this possible and quickly impacted many problems in computer vision (see reviews~\cite{SCHMIDHUBER201585, litjens2017survey, serre2019deep}).

\begin{figure*}
\begin{center}
\includegraphics[width=\textwidth]{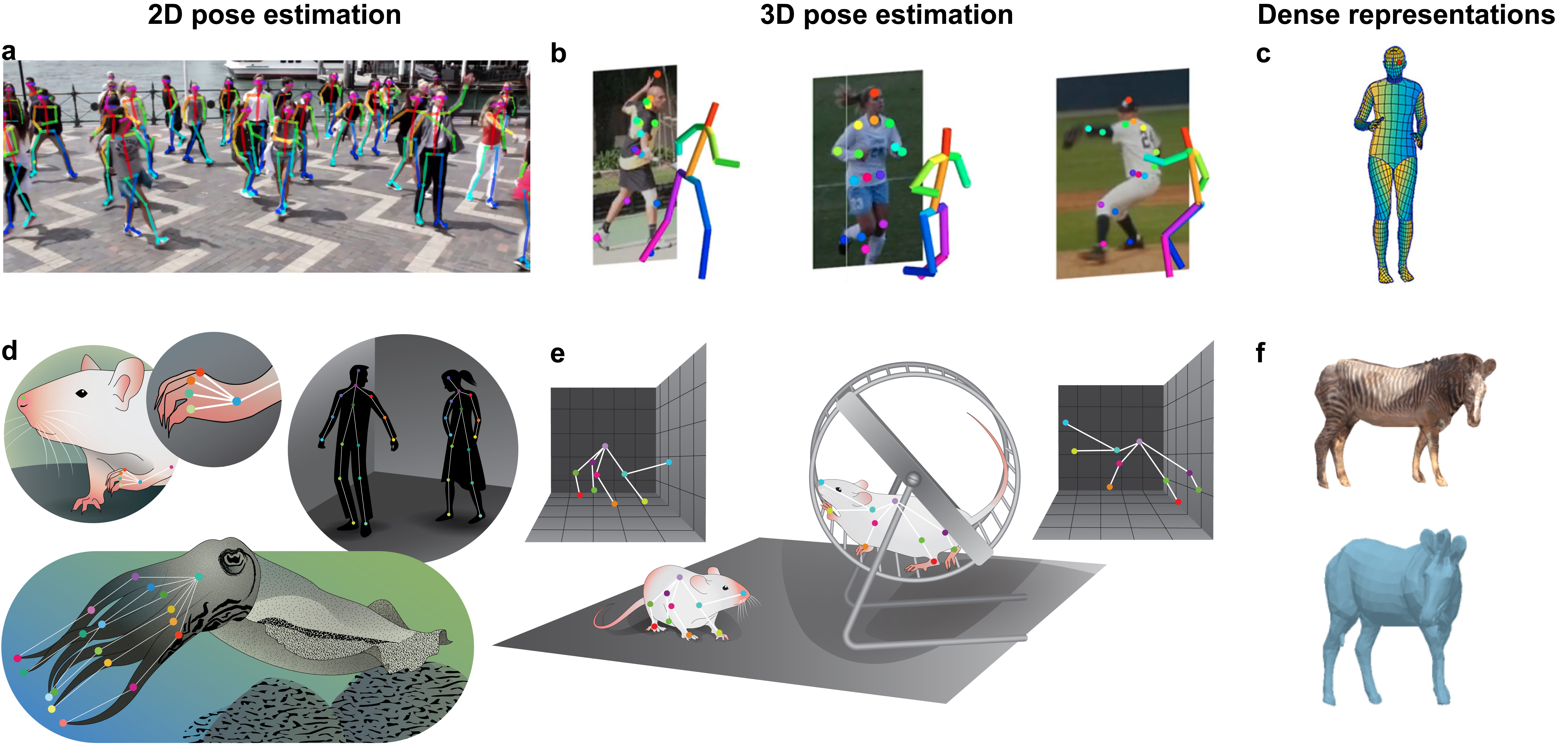}
\end{center}
\caption{{\bf 2D pose estimation, 3D pose estimation \& dense representations of humans and other animals:} {\bf a:} Example 2D multi-human pose estimation from OpenPose~\cite{cao2017realtime}. {\bf b:} Example 3D human pose estimation from~\citep{Mehta2017_3D}. {\bf c:} Dense representations of humans with DensePose, adapted from Guler et al.~\cite{guler2018densepose}. {\bf d:} Animals have diverse bodies and experimenter's are often interested in specific key points, making tailored network attractive solutions. DNNs open a realm of possibilities: from mice to cuttlefish. {\bf e:} 3D pose estimation requires multiple cameras views, or 2D to 3D ``lifting". {\bf f:} The new SMALST model which fits full 3D models to images from Zuffi et al.~\citep{Zuffi2019ICCV} applied to zebras.}
\label{fig:2D3D}
\end{figure*}

\subsection*{2D and 3D (human) pose estimation}

\justify In 2014 ``DeepPose" was the first paper to apply deep learning to human 2D pose estimation~\cite{toshev2014deeppose}, and immediately new networks were proposed that improved accuracy by introducing a translation invariant model~\cite{JainMODEEP}, and convolutional networks plus geometric constraints~\cite{NIPS2014_5573, Tompson_2015_CVPR}. In the few years since, numerous human pose estimation papers (approx. 4,000 on Google Scholar), and new benchmarks with standardized datasets plus evaluation metrics appeared, which allow better comparisons of ``state-of-the-art" performance~\cite{andriluka2018posetrack}. This culture has driven rapid and remarkable increases in performance: from ~44\% of body parts correctly labeled to nearly 94\% - with the top 15 networks being within a few percentage points of each other (an example top network is shown in Figure~\ref{fig:2D3D}a) ~\cite{newell2016stacked, insafutdinov2016deepercut, insafutdinov2017cvpr, cao2017realtime, kreiss2019pifpaf, sun2019deep}. The history and many advances in 2D human pose estimation are comprehensively reviewed in~\cite{POPPE20074, Dang2019}.\\

3D human pose estimation is a more challenging task and 3D labeled data is more difficult to acquire. There has been massive improvements in networks; see review~\cite{Sarafianos2016}. Yet currently, the highest accuracy is achieved by using multiple 2D views to reconstruct a 3D estimate (Figure~\ref{fig:2D3D}b;~\cite{martinez2017simple, Mehta2017_3D}), but other ways of ``lifting" 2D into 3D are being actively explored~\cite{TomeRA17,chen20173d,Mehta2017_3D}.

\subsection*{Dense-representations of bodies}

\justify Other video-based approaches for capturing the posture and soft tissue of humans (and other animals) also exist. Depth-cameras such as the Microsoft Kinect have been used in humans~\cite{Shotton2012EfficientHP, kinectEval2012} and rodents~\cite{wiltschko2015mapping, hong2015automated}. Recently dense-pose representations, i.e. 3D point clouds or meshes (Figure~\ref{fig:2D3D}c), have become a popular and elegant way to capture the soft-tissue and shape of bodies, which are highly important features for person identification, fashion (i.e. clothing sales), and in medicine~\cite{guler2018densepose, SMPL-X:2019, kanazawaHMR18}. However, state-of-the-art performance currently requires body-scanning of many subjects to make body models. Typically, large datasets are collected to enable the creation of robust algorithms for inference on diverse humans (or for animals, scanning toy models has been fruitful~\cite{Zuffi20163dMenagerie}). Recently, outstanding improvements have been made to capture shapes of animals from images~\cite{biggs2018creatures, Zuffi_2018_CVPR, Zuffi2019ICCV}. However, there are no animal-specific toolboxes geared towards neuroscience applications, although we believe that this will change in the near future, as for many applications having the soft-tissue measured will be highly important, i.e. in obesity or pregnancy research, etc. 

\section*{Animal pose estimation}

The remarkable performance when using deep learning for human 2D \& 3D pose estimation plus dense-representations made this large body of work ripe for exploring its utility in neuroscience (Figure~\ref{fig:2D3D}d-f). In the past two years, deep learning tools for laboratory experiments have arrived (Figure~\ref{fig:examples}a-d).\\

Many of the properties of DNNs were extremely appealing: remarkable and robust performance, relatively fast inference due to GPU hardware, and efficient code due to modern packages like TensorFlow and PyTorch (reviewed in~\cite{Nguyen2019}). Furthermore, unlike for many previous algorithms, neither body models nor tedious manual tuning of parameters is required. Given the algorithms, the crucial ingredient for human pose estimation success was large-scale well annotated data sets of humans with the locations of the bodyparts. \\

Here, we identify {\bf 5} key areas that were important for making DNN-based pose estimation tools useful for neuroscience laboratories, and review the progress in the last two years:

\begin{enumerate}
\item Can DNNs be harnessed with small training datasets? Due to the nature of ``small-scale" laboratory experiments, labeling $>20,000$ or more frames is not a feasible approach (the typical human benchmark dataset sizes). 
\item The end-result must be as accurate as a human manually-applied labels (i.e. the gold standard), and computationally tractable (fast).
\item The resulting networks should be robust to changes in experimental setups, and for long-term storage and re-analysis of video data, to video compression. 
\item Animals move in 3D, thus having efficient solutions for 3D pose estimation would be highly valuable, especially in the context of studying motor learning and control. 
\item Tracking multiple subjects and objects is important for many experiments studying social behaviors as well as for animal-object interactions.
\end{enumerate}

\subsection*{1. Small training sets for lab-sized experiments}

\justify While the challenges discussed above for human pose estimation also apply for other animals, one important challenge for applying these methods to neuroscience was annotated data sets - could DNNs be harnessed for much smaller datasets, at sizes reasonable for typical labs? Thus, while it was clear that given enough annotated frames the same algorithms will be able to learn to track the body parts of any animal, there were feasibility concerns.\\

Human networks are typically trained on thousands of images, and nearly all the current state-of-the-art networks provide tailored solutions that utilize the skeleton structure during inference~\cite{cao2017realtime, insafutdinov2017cvpr}. Thus, applying these tools to new datasets was not immediately straight-forward, and to create animal-specific networks one would need to potentially curate large datasets of the animal(s) they wanted to track. Additionally, researchers would need tailored DNNs to track their subjects (plus the ability to track unique objects, such as the corners of a box, or an implanted fiber).\\

Thus, one of the most important challenges is creating tailored DNNs that are robust and generalize well with little training data. One potential solution for making networks for animal pose estimation that could generalize well, even with little data, was to use transfer learning - the ability to take a network that has been trained on one task to perform another. The advantage is that these networks are pretrained on larger datasets (for different tasks where a lot of data is available like ImageNet), therefore they are effectively imbued with good image representations.\\
 
This is indeed what ``DeepLabCut," the first tool to leverage the advances in human pose estimation for application to animals did \cite{mathis2018deeplabcut}. DeepLabCut was built on a subset of ``DeeperCut"~\cite{insafutdinov2016deepercut}, 
which was an attractive option due to its use of ResNets, which are powerful for transfer learning~\cite{kornblith2019better, mathis2019TRANSFER}. Moreover transfer learning reduces training times~\cite{kornblith2019better, mathis2019TRANSFER, arac2019deepbehavior}, and there is a significant gain over using randomly-initialized networks in performance, especially for smaller datasets~\cite{mathis2019TRANSFER}.\\

\begin{figure*}
\begin{center}
\includegraphics[width=.85\textwidth]{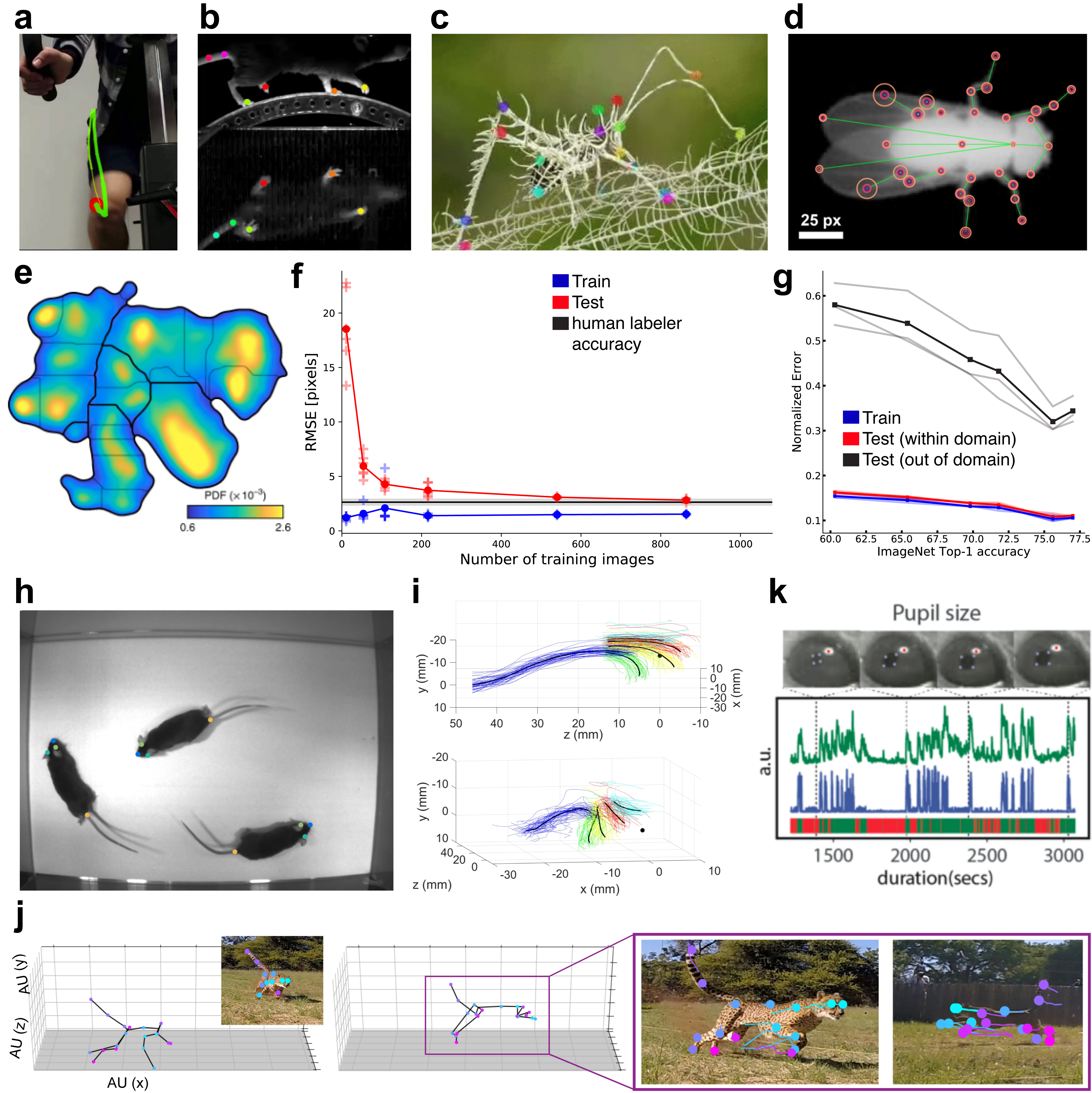}
\end{center}
\caption{{\bf DNNs applied to animal pose estimation.} {\bf a:} Knee tracking during cycling adopted from~\cite{kaplan2019video}. {\bf b:} 3D limb estimates from~\cite{MathisWarren2018speed}. {\bf c:} A Lichen Katydid tracked with DeepLabCut, courtesy of the authors. {\bf d:} Fly with LEAP annotated body parts. The circles indicate the fraction for which predicted positions of the particular body part are closer to the ground truth than the radii on test images (adopted from~\cite{pereira2019fast}). {\bf e:} Density plot of t-SNE plus frequency-transformed freely moving fly body-part trajectories. Patches with higher probability indicate more common movements like different types of grooming behaviors (adopted from~\cite{pereira2019fast}); {\bf f.} DeepLabCut requires little data to match human performance. Average error (RMSE) for several splits of training and test data vs. number of training images compared to RMSE of a human scorer. Each split is denoted by a cross, the average by a dot. For $80\%$ of the data, the algorithm achieves human level accuracy on the test set. As expected, test RMSE increases for fewer training images. Around 100 diverse frames are enough to provide high tracking performance (<5-pixel accuracy - adopted from~\cite{mathis2018deeplabcut}). {\bf g:} Networks that perform better on ImageNet perform better for predicting 22 body parts on horses on within-domain (similar data distribution as training set, red) and out-of-domain data (novel horses, black). The faint lines are individual splits. (adopted from~\cite{mathis2019TRANSFER}). {\bf h:} Due to the convolutional network architecture, when trained on one mouse the network generalizes to detect body parts of three mice (adopted from~\cite{mathis2018deeplabcut}). {\bf i:} 3D reaching kinematics of rat (adopted from~\cite{bova2019automated}). {\bf j:} 3D pose estimation on a cheetah for 2 example poses from 6 cameras as well es example 2D views (adopted from~\cite{nath2019deeplabcut}). {\bf k:} Pupil and pupil-size tracking (adopted from~\cite{sriram2019sparse}).}
\label{fig:examples}
\end{figure*}

The major result from DeepLabCut was benchmarking on smaller datasets and finding that only a few hundred annotated images are enough to achieve excellent results for diverse pose estimation tasks like locomotion, reaching and trail-tracking in mice, egg-laying in flies and hunting in cheetahs, due to transfer learning (Figure~\ref{fig:examples}f,g,h)~\cite{MathisWarren2018speed, nath2019deeplabcut, mathis2019TRANSFER}. ``DeepBehavior," which utilized different DNN-packages for various pose estimation problems, also illustrated the gain of transfer learning~\cite{arac2019deepbehavior}.
 
\subsection*{2. Accuracy \& speed}

\justify  To be useful, pose estimation tools need to be as good as human annotation of frames (or tracking markers, another proxy for a human-applied label). DeepLabCut was shown to reach human labeling accuracy~\cite{mathis2018deeplabcut}, and can replace physical markers~\cite{moore2019}. Moreover, they need to be efficient (fast) for both offline analysis and online analysis. Speed is often related to the depth of the network. Stacked-hourglass networks, which use iterative refinement~\cite{newell2016stacked,SegNet} and fewer layers, are fast. Two toolboxes, ``LEAP" ~\cite{pereira2019fast} and ``DeepPoseKit"~\cite{graving2019fast} adopted variants of stacked-hourglass networks. LEAP allows the user to rapidly compute postures, and then perform unsupervised behavioral analysis (Figure~\ref{fig:examples}d,e)~\cite{Berman2014}. This is an attractive solution for real-time applications, but it is not quite as accurate. For various datasets, DeepPoseKit reports it is about three times as accurate as LEAP, yet similar to DeepLabCut~\cite{graving2019fast}. They also report about twice faster video processing compared to DeepLabCut and LEAP for batch-processing (on small frame sizes).\\

Deeper networks are slower, but often have more generalization ability~\cite{kornblith2019better}. DeepLabCut was designed for generalization and therefore utilized deeper networks (ResNets) that are inherently slower than stacked-hourglass networks, yet DeepLabCut can match human accuracy in labeling (Figure~\ref{fig:examples}f)~\cite{mathis2018deeplabcut}. The speed has been shown to be compatible with online-feedback applications~\cite{forys2018real,vstih2019stytra,MathisWarren2018speed}. Other networks recently added to DeepLabCut (with a MobileNetV2 backbone) give slightly lower accuracy, with twice the speed~\cite{mathis2019TRANSFER}. Overall, on GPU hardware all packages are fast, and reach speeds of several hundred frames per second in offline modes.

\subsection*{3. Robustness}

\justify Neuroscience experiments based on video recordings produce large quantities of data and are collected over extensive periods of time. Thus, analysis pipelines should be robust to a myriad of perturbations: such as changes in setups (backgrounds, light sources, cameras, etc.), subject appearance (due to different animal strains), and compression algorithms (which allow storage of perceptually good videos with little memory demands~\cite{wiegand2003overview}).

How can robustness be increased within the DNN? Both transfer learning (discussed above) and data augmentation strategies are popular and rapidly evolving approaches to increase robustness in DNNs (see review~\cite{Shorten2019}). Moreover, active learning approaches allow an experimenter to continuously build more robust and diverse datasets, for large scale projects by expanding the training set with images, where the network fails~\cite{mathis2018deeplabcut, nath2019deeplabcut, pereira2019fast}. So far, the toolboxes have been tested on data from the same distribution (i.e. by splitting frames from videos into test and training data), which is important for assessing the performance~\cite{pereira2019fast,graving2019fast,mathis2018deeplabcut}, but did not directly tested out-of-domain robustness.\\

Over the course of long-term experiments the background or even animal strain can change, which means having robust networks would be highly advantageous. We recently tested the generalization ability of DeepLabCut with different network backbones for pose estimation. We find that pretraining on ImageNet strongly improves out-of-domain performance, and that better ImageNet performing networks are more robust (Figure~\ref{fig:examples}g)~\cite{mathis2019TRANSFER}. There is still a gap to close in out-of-domain performance, however.\\

DeepLabCut is also robust to video compression, as compression by more than 1,000X only mildly affects accuracy (less than 1 pixel average error less)~\cite{MathisWarren2018speed}. The International Brain Lab (IBL) independently and synergistically showed that for tracking multiple body parts in a rodent decision making task, DeepLabCut is robust to video compression~\cite{Meijer2019}. Thus, in practice users can substantially compress videos, while retaining accurate posture information.

\begin{figure*}
\begin{center}
\includegraphics[width=.97\textwidth]{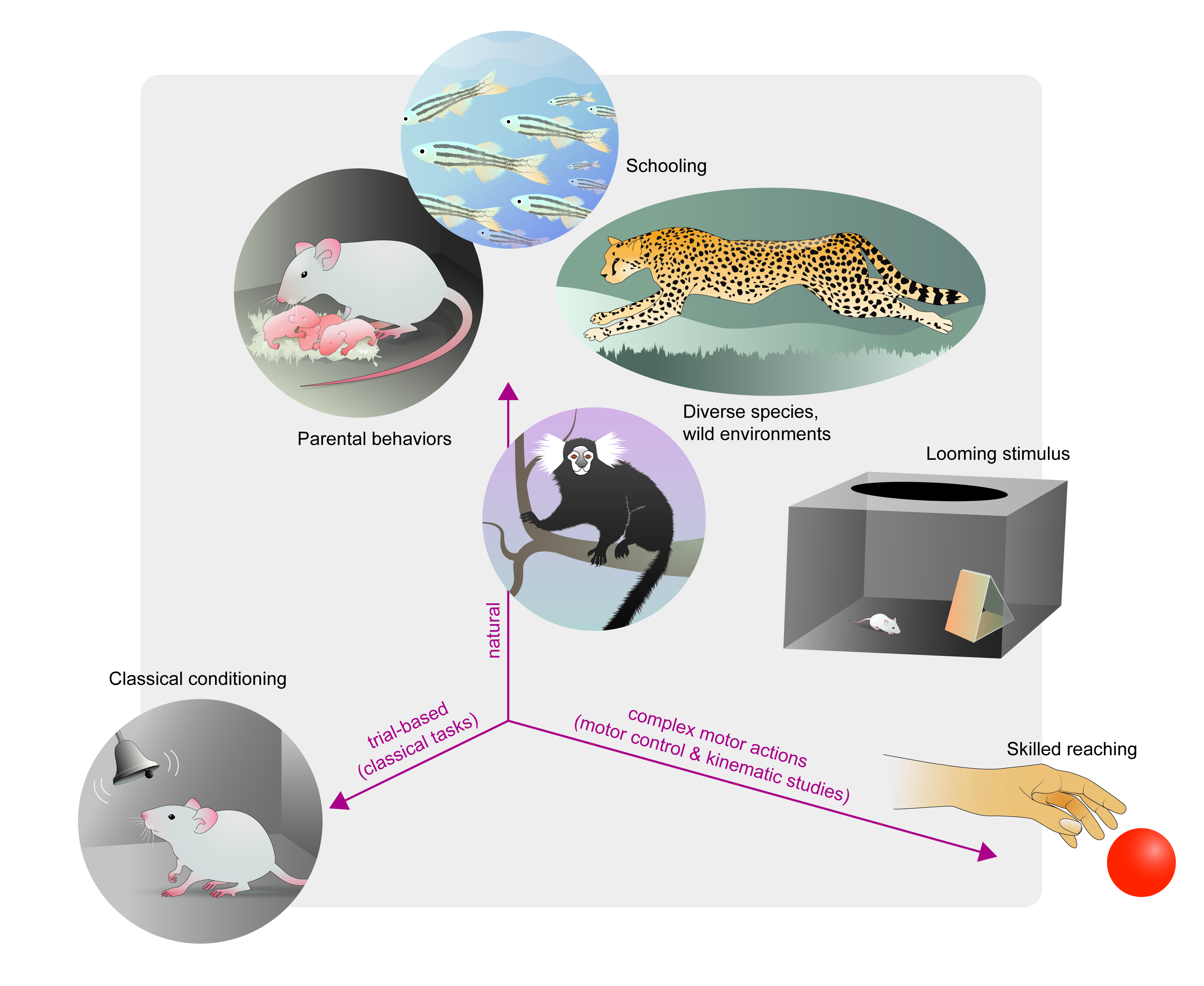}
\end{center}
\caption{{\bf The Behavioral Space in Neuroscience: new applications for deep learning-assisted analysis.} This diagram depicts how pose estimation with non-invasive videography can benefit behavioral paradigms that span from ``trial-based behaviors" such as classical conditioning, to ``complex motor behaviors/tasks" as in 3D reaching assays, to ``naturalistic tasks," often without any trial structure, and that are more akin to real-world 'tasks' that animals undertake. For example, a looming stimulus is ethologically relevant and complex, and pup-retrieval can be natural yet repeatable in a laboratory setting. With new tools that allow for fast and accurate analysis of movement, these types of experiments become more feasible (with much less human labor than previously required).}
\label{fig:toolbox}
\end{figure*}

\subsection*{4. 3D animal pose estimation}

\justify Currently, there are several animal pose estimation toolboxes that explicitly support 2D and 3D key-point detection~\cite{nath2019deeplabcut, pierre2019, pavan2019, zhang2019multiview}. DeepLabCut uses 2D pose estimation to train a single camera-invariant 2D network (or multiple 2D networks) that is then used to perform traditional triangulation to extract 3D key points (Figure~\ref{fig:examples}i, j; ~\cite{nath2019deeplabcut, bova2019automated}). A pipeline built on DeepLabCut called ``Anipose" allows for 3D reconstruction from multiple cameras using a wider variety of methods~\cite{pierre2019}. ``DeepFly3D"~\cite{pavan2019} uses the network architecture from Newell et al.~\cite{newell2016stacked} and then adds elegant tools to compute an accurate 3D estimate of \textit{Drosophila melanogaster} by using the fly itself vs. standard calibration boards. Zhang et al. use epipolar geometry to train across views and thereby improve 3D pose estimation for mice, dogs, and monkeys~\cite{zhang2019multiview}. 

\subsection*{5. Multi-animal \& object tracking}

\justify Many experiments in neuroscience require measuring interactions of multiple animals or interactions with objects. Having the ability to both flexibly track user-defined objects or multiple animals therefore is highly desirable. There are many pre-deep learning algorithms that allow tracking of objects (one such modern example called ``Tracktor" also nicely summarizes this body of work~\citep{Sridhar2019}). Recently researchers have also applied deep learning to this problem. For example, the impressive  ``idTracker:ai"~\cite{romero2019idtracker} allows for users to track a hundred individual, unmarked animals. Arac et al. used YOLO, a popular and fast object localization network, for tracking two mice during a social behavior~\cite{arac2019deepbehavior}. These, and others, can then be combined with pose estimation packages for estimating the pose of multiple animals. Currently, two paths are possible: one is to apply pose estimation algorithms after tracking individuals (for which any package could be used); or, two, extract multiple detections for each part on each animal (Figure~\ref{fig:examples}h;~\cite{mathis2018deeplabcut, jiang2019detection}) and link them using part affinity fields~\cite{cao2017realtime}, pairwise predictions~\cite{insafutdinov2016deepercut}, or geometrical constraints~\cite{jiang2019detection}, plus combinatorics.

\section*{The impact on experimental neuroscience}

In the short time period these tools have become available there has been a rather wide adoption by the neuroscience and ethology community. Beyond the original publications, DeepLabCut has already been used for pose estimation and behavioral analysis in many contexts. For instance, knee movement quantification during cycling (Figure~\ref{fig:examples}a)~\cite{kaplan2019video},  postural analysis during underwater running~\cite{cronin2019markerless}, social behavior in bats~\cite{zhang2019correlated}, for benchmarking thermal constraints with optogenetics~\cite{owen2019thermal}, fly leg movement analysis~\cite{bidaye2019two, azevedo2019size}, for 3D rat reaching (Figure~\ref{fig:examples}i)~\cite{bova2019automated}, hydra in a thermal stimulation assay~\cite{tzouanas2019thermal} and pupillometry (Figure~\ref{fig:examples}k) ~\cite{sriram2019sparse}. Also inanimate objects can be tracked, and it has indeed also been used to track metal beads when subjected to a high voltage~\cite{de2019oscillatory}, and magic tricks (i.e. coins and the magician)~\cite{zaghi2019playing}. LEAP~\cite{pereira2019fast} has been used to track ants~\cite{clifton2019rough} and mice~\cite{Yue2019_leapmice}.\\

Pose estimation is just the beginning; the next steps involve careful analysis of kinematics, building detailed, multi-scale ethograms of behaviors, new modeling techniques to understand large-scale brain activity and behaviors across a multitude of timescales, and beyond. We envision three branches where powerful feature tracking and extensions will be useful: motor control studies (often involving complex motor actions), naturalistic behaviors in the lab and in the wild, and better quantification of robust and seemingly simple ``non-motor" tasks (Figure~\ref{fig:toolbox}).\\

Many paradigms in neuroscience can be loosely arranged along three branches as natural (i.e. mouse parenting behavior), simple trail-based tasks (i.e. classical conditioning), and/or complex motor actions like skilled reaching (Figure~\ref{fig:toolbox}). For example, you can have simple and natural tasks such as licking for water, or complex and natural behaviors such as escaping from a looming stimulus that would rarely produce repeated trajectories. For simplicity, here we discuss how pose estimation can potentially enhance these studies along those three branches, namely complex movements (Motor control \& kinematics), natural behaviors (Natural behaviors \& ethologically relevant features), and during simple motor-output tasks (Revisiting classic tasks).

\subsection*{Motor control \& kinematics}

\justify Often in neuroscience-minded motor control studies end-effector proxies (such as manipulandums or joysticks) are used to measure the motor behavior of subjects or animals. There are relatively few marker-tracking based movement neuroscience studies, in which many degrees of freedom were measured alongside neural activity, with notable exceptions like~\cite{vargas2010decoding, schaffelhofer2015decoding}. The ease with which kinematics of limbs and digits can now be quantified~\cite{mathis2018deeplabcut, bova2019automated, azevedo2019size, arac2019deepbehavior} should greatly simplify such studies in the future. We expect many more highly detailed kinematic studies will emerge that utilize DNN-based analyses, especially for freely moving animals, for small and aquatic animals that cannot be marked, and for motor control studies that can leverage large-scale recordings and behavioral monitoring.

\subsection*{Natural behaviors \& ethologically relevant features}

\justify There is a trend in motor neuroscience towards natural behaviors; i.e. less constrained tasks, everyday-skills, and even ``in the wild" studies~\cite{mathis2019highlights}. For instance, we used DeepLabCut for 3D pose estimation in hunting cheetah's captured via multiple Go-Pro cameras (Figure~\ref{fig:examples}j;~\cite{nath2019deeplabcut}). Another ``in the wild example" is given by a recent study by Chambers et al.~\cite{chambers2019pose}, who revisited the classic question of how people synchronize their walking, but with a modern twist by using videos from YouTube and analysis with OpenPose~\cite{cao2017realtime}. Consistent with studies performed in the laboratory, they found a tendency for pairs of people to either walk in or exactly out of phase~\cite{chambers2019pose}.\\

How else can DNNs help? Specialized body parts often play a key role in ethologically relevant behaviors. For instance, ants use their antenna to follow odor trails~\cite{draft2018carpenter}, while moles use their snouts for sampling bilateral nasal cues to localize odorants~\cite{catania2013stereo}. To accurately measure such behaviors, highly accurate feature-detectors of often tiny, highly dexterous bodyparts are needed. This is a situation where deep learning algorithms can excel. Pose estimation algorithms can not only be used to detect the complete "pose", but due to their flexibility they are extremely useful to track ethologically relevant body parts in challenging situations; incidentally DeepLabCut was created, in part, to accurately track the snouts of mice following odor trails that were printed onto a treadmill~\cite{mathis2018deeplabcut}. There are of course many other specialized body parts that are hard to track: like whiskers, bee-stingers, jellyfish tentacles, or octopus arms, and we believe that studying these beautiful systems in more natural and ethologically relevant environments has now gotten easier. 

\subsection*{Revisiting classic tasks}

\justify Measuring behavior is already impacting ``classic" decision-making paradigms. For example, several groups could show broad movement encoding across the brain during decision-making tasks by carefully quantifying behavior~\cite{Stringereaav7893, Musall_NN}. Moreover, large scale efforts to use these ``simple" yet robust trial-based behaviors across labs and brain areas are leveraging deep learning, and comparing their utility compared to classical behavior-monitoring approaches. For example, the IBL has surmised that DeepLabCut could replace traditional methods used for eye, paw and lick detection~\cite{Meijer2019}. We believe that detailed behavioral analysis will impact many paradigms, which were historically not considered ``motor" studies, as now it is much easier to measure movement. 

\subsection*{Remaining challenges in pose estimation}

\justify Advances in deep learning have changed how easily posture can be measured and has impacted many studies. However, pose estimation remains a hard computer vision problem and challenges remain~\cite{andriluka2018posetrack, golda2019human, Dang2019, serre2019deep, rhodin2018learning, kordingLimitations2019}. In the context of multi-animal/human pose estimation, dealing with highly crowded scenes, in which different individuals cover each other, remains highly challenging~\cite{andriluka2018posetrack, Dang2019, golda2019human}. In general, dealing with occlusions remains a challenge. In some experiments occlusions are hard to avoid. Thus, networks that can constraint body part detection based on anatomical relationships can be advantageous, but are computationally more complex and slower~\cite{insafutdinov2016deepercut}.
As we highlighted in the robustness section, it is hard to train networks to generalize to out-of-domain scenarios~\cite{rhodin2018learning, mathis2019TRANSFER, michaelis2019benchmarking}. Even though very large data sets have been amassed to build robust DNNs~\cite{insafutdinov2017cvpr, cao2017realtime, kreiss2019pifpaf, sun2019deep}, they can still fail in sufficiently different scenarios~\cite{rhodin2018learning, mathis2019TRANSFER, kordingLimitations2019}. Making robust networks will highly useful for creating shareable behavior- or animal-specific networks that can generalize across laboratories. There will also be much work towards even faster, lighter models.

\section*{Outlook \& Conclusions}

\justify The field of 2D, 3D, and dense pose estimation will continue to evolve. For example, with respect to handling occlusions and robustness to out-of-domain data. Perhaps larger and more balanced datasets will be created to better span the behavioral space, more temporal information will be utilized when training networks or analyzing data, and new algorithmic solutions will be found.\\

Will we always need training data? A hot topic in object recognition is training from very few examples (one-shot or zero-shot learning)~\cite{Xian2019_ZERO}. Can this be achieved in pose estimation? Perhaps as new architectures and training regimes come to fruition this could be possible. Alternatively, specialized networks could now be built that leverage large datasets of specific animals. It is hard to envision a universal ``animal pose detector" network (for object recognition this is possible) as animals have highly diverse body plans and experimentalists often have extremely different needs. Currently many individual labs create their own specialized networks, but we plan to create shareable networks for specific animals (much like the specific networks, i.e. hand network in OpenPose~\cite{simon2017hand}, or the human-network in DeepLabCut~\cite{insafutdinov2016deepercut, mathis2018deeplabcut}). For example, many open field experiments could benefit from robust and easy-to-use DNNs for video analysis across similar body points of the mouse. Indeed, efforts are underway to create networks where one can simply analyze their data without training, and we hope the community will join these efforts. Nothing improves DNNs more than more training data. These efforts, together with making code open source, will contribute to the reproducibility of science and make these tools broadly accessible.\\

{\bf In summary,} we aimed to review the progress in computer vision for human pose estimation, how it influenced animal pose estimation, and how neuroscience laboratories can leverage these tools for better quantification of behavior. Taken together, the tremendous advance of computer vision has provided tools that are practical for the use in the laboratory, and they will only get better. They can be as accurate as human-labeling (or marker-based tracking), and are fast enough for closed-loop experiments, which is key for understanding the link between neural systems and behavior. We expect that in-light of shared, easy-to-use tools and additional deep learning advances, there will be thrilling and unforeseen advances in real-world neuroscience. 

\begin{flushleft}
\textbf{Acknowledgments:} 
\end{flushleft}
We thank Eldar Insafutdinov, Alex Gomez-Marin, the M. Mathis Lab, and the anonymous reviewers for comments on the manuscript, and Julia Kuhl for illustrations. Funding was provided by the Rowland Institute at Harvard University (M.W.M.) and NIH U19MH114823 (A.M). The authors declare no conflicts of interest.

\section*{References}
%\bibliographystyle{unsrt}
%\bibliography{literature}

\section*{Highlighted References:}

[1*] {\bf A survey on deep learning in medical image analysis}~\cite{litjens2017survey}
\textit{Comprehensive review of all deep learning algorithms used in medical image analysis as of 2017, as well as a discussion of most successful approaches, together with [2**] a fantastic introduction for newcomers to the field.}\\

[2**] {\bf Deep learning: The good, the bad, and the ugly}~\cite{serre2019deep}
\textit{Excellent review of deep learning progress including a detailed description of recent successes as well as limitations of computer vision algorithms.}\\

[3**] {\bf Realtime multi-person 2d pose estimation using part affinity fields}~\citep{cao2017realtime}
\textit{OpenPose was the first real-time multi-person system to jointly detect human body parts by using part affinity fields, a great way to link body part proposals across individuals. The toolbox is well maintained and now boasts body, hand, facial, and foot keypoints (in total 135 keypoints) as well as 3D support.}
\newpage

[4*] {\bf Dense-pose: Dense human pose estimation in the wild.}\cite{guler2018densepose}
\textit{Using a large dataset of humans (50K), they build dense correspondences between RGB images and human bodies. They apply this to human ``in the wild," and build tools for efficiently dealing with occlusions. It is highly accurate and runs up to 25 frames per second.}\\

[5**] {\bf Three-D Safari: Learning to Estimate Zebra Pose, Shape, and Texture from Images ``In the Wild"}~\citep{Zuffi2019ICCV}
\textit{Zuffi et al. push dense pose estimations by using a new SMALST model for capturing zebras pose, soft-tissue shape, and even texture ``in the wild." This is a difficult challenge as zebras are designed to blend into the background in the safari. This paper makes significant improvements on accuracy and realism, and builds on a line of elegant work from these authors~\cite{Zuffi20163dMenagerie, Zuffi_2018_CVPR}}\\

[6**] {\bf DeeperCut: A deeper, stronger, and faster multi-person pose estimation model}~\citep{insafutdinov2016deepercut}
\textit{DeeperCut is a highly accurate algorithm for multi-human pose estimation due to improved deep learning based body part detectors, and image-conditioned pairwise terms to predict the location of body parts based on the location of other body parts. These terms are then used to find accurate poses of individuals via graph cutting. In ArtTrack~\cite{insafutdinov2017cvpr} the work was extended to fast multi-human pose estimation in videos.}\\

[7*] {\bf Recovering the shape and motion of animals from video}~\cite{biggs2018creatures}
\textit{The authors combine multiple methods in order to efficiently fit 3D shape to multiple quadrupeds from camels to bears. They also provides a novel dataset of joint annotations and silhouette segmentation for eleven animal videos.}\\
\vfill\eject

[8**] {\bf DeepLabCut: markerless pose estimation of user-defined body parts with deep learning}~\citep{mathis2018deeplabcut}
\textit{DeepLabCut was the first deep learning toolbox for animal pose estimation. The key advance was to benchmark a subset of the feature detectors in  DeeperCut~\cite{insafutdinov2017cvpr}. This paper showed nearly human-level performance with only 50-200 images. It benchmarked flies moving in a 3D chamber, hand articulations and open-field behavior in mice, and provided open-source tools for creating new datasets and data loaders to train deep neural networks, and post-hoc analysis. Subsequent work has improved accuracy, speed, and introduced more network variants into the Python package~\cite{MathisWarren2018speed, nath2019deeplabcut, mathis2019TRANSFER}.}\\

[9*] {\bf DeepBehavior: A deep learning toolbox for automated analysis of animal and human behavior imaging data}~\cite{arac2019deepbehavior}
\textit{Arac et al. use three different DNN-packages (OpenPose~\cite{cao2017realtime}, YOLO and Tensorbox) for analyzing 3D analysis of pellet reaching, three-chamber test, social behavior in mice, and 3D human kinematics analysis for clinical motor function assessment with OpenPose~\cite{cao2017realtime}. They also provide MATLAB scripts for additional analysis (after pose estimation).}\\

[10*] {\bf Using DeepLabCut for 3D markerless pose estimation across species and behaviors}~\cite{nath2019deeplabcut}. 
\textit{A Nature Protocols user-guide to DeepLabCut2.0, with 3D pose estimation of hunting cheetahs and improved network performance. The toolbox is provided as a Python package with graphical user interfaces for labeling, active-learning-based network refinement, together with Jupyter Notebooks that can be run on cloud resources such as Google Colaboratory (for free).}\\

[11*] {\bf Fast  animal  pose  estimation  using  deep  neural networks}~\cite{pereira2019fast}. 
\textit{LEAP (LEAP estimates animal pose), a DNN method for predicting the positions of animal body parts. This framework consists of a graphical interface for labeling of body parts and training the network. Training and inference times are fast due to the lightweight architecture. The authors also analyzed insect gait based on unsupervised behavioral methods~\cite{Berman2014}, which they directly applied to the posture, rather than to compressed image features.}

\end{document}